# IMPLEMENTATION OF AN INDEX OPTIMIZE TECHNOLOGY FOR HIGHLY SPECIALIZED TERMS BASED ON THE PHONETIC ALGORITHM METAPHONE

*При формуванні баз даних, наприклад для задоволення потреб закладів охорони здоров'я, доволі часто виникає проблема щодо введення та подальшої обробки імен і прізвищ лікарів і пацієнтів, які є вузькоспеціалізованими за вимовою і написанням. Це пояснюється тим, що імена та прізвища людей не можуть бути унікальними, їх напис не підпадає під жодні правила фонетики, а їх довжини при їх викладенні різними мовами можуть не співпадати. З появою інтернету такий стан справ стає взагалі критичним й може привести до того, що за однією адресою може бути відправлено декілька копій електронних листів. Вирішити означену проблему можуть допомогти фонетичні алгоритми порівняння слів Daitch-Mokotoff, Soundex, NYSIIS, Polyphone та Metaphone, а також алгоритми Левенштейна та Джаро, алгоритми на основі Q-грам, які дозволяють знаходити відстані між словами. Найбільшого поширення серед них отримали алгоритми Soundex і Metaphone, які призначені для індексування слів по їх звучанням з урахуванням правил вимови. Шляхом застосування алгоритму Metaphone зроблено спробу оптимізації процесів фонетичного пошуку для задач нечіткого співпадіння, наприклад, при дедублікації даних в різноманітних базах даних і реєстрах для зменшення кількості помилок невірного введення прізвищ. Із аналізу найбільш розповсюджених прізвищ видно, що частина з них є українського або російського походження. При цьому правила, за якими вимовляються і записуються прізвища, наприклад українською мовою, кардинально відрізняються від базових алгоритмів для англійської і достатньо відрізняються для російської мови. Саме тому фонетичний алгоритм має враховувати передусім особливості формування українських прізвищ, що нині є надзвичайно актуальним. Представлено результати експерименту із формування фонетичних індексів, а також результати збільшення продуктивності при використанні сформованих індексів. Окремо представлено метод адаптації пошуку для інших сфер і кількох споріднених мов на прикладі пошуку по лікарським засобам*

*Ключові слова: нечітке співпадіння, фонетичне правило, фонетичний алгоритм, Metaphone, українське прізвище*

**V. Buriachok**
Doctor of Technical Sciences, Professor*

**M. Hadzhyiev**
Doctor of Technical Sciences, Associate Professor
Department of Information Security and Data Transfer
Odessa National O. S. Popov Academy of Telecommunications
Kuznechna str., 1, Odessa, Ukraine, 65029

**V. Sokolov**
Postgraduate Student*

**P. Skladannyi**
Postgraduate Student*
E-mail: p.skladannyi@kubg.edu.ua

**L. Kuzmenko**
Postgraduate Student
Institute of Telecommunications and Global Information Space of the National Academy of Sciences of Ukraine
Chokolivskiy blvd., 13, Kyiv, Ukraine, 03186
*Department of Information and Cyber Security
Borys Grinchenko Kyiv University
Bulvarno-Kudriavska str., 18/2, Kyiv, Ukraine, 04053

## 1. Introduction

A variety of mechanisms and approaches can be used to search for fuzzy matches between words and phrases: distance calculation by Levenshtein, Damerau-Levenshtein, or Hemming, similarities by Jaro or Jaro-Winkler, construction of Q-grams, etc. [1]. These algorithms are universal and their use is justified when analyzing long literals with finite alphabets (including an analysis of similarity and search for mutations in DNA and RNA). In addition, phonetic algorithms can be used to define *n*-multiple errors in typos and misspellings, but none of them:

– firstly, does not take into consideration the peculiarities of sound perception of unfamiliar words (in this case, last names) by human;

– secondly, cannot fully define determining phonetic errors (such errors occur in languages with old grammar and caused by cancellation of or variability in pronunciation and written reproduction).

That is why there is a task to simplify and unify the process of sound perception. This is primarily predetermined by the medical reform, within which the automation of medical records is conducted by heads of hospitals, doctors, pharmacists, laboratory staff, diagnostic and junior medical personnel, as well as by patients. Transferring data into digital format leads, in turn, to new challenges: operating the highly specialized terms – personal data on patients and titles of medical preparations. Given the above, it is an urgent task to improve, first of all, medical information systems at the stages of entering new data and searching through existing databases.

## 2. Literature review and problem statement

The task of searching for highly specialized words/terms is reduced, as is known, to the formation of search indexes based on a phonetic key. Currently, this task has to a certain degree been formalized for English [2, 3], Russian [4, 5] and some other languages. To this end, there are several phonetic algorithms and their modifications that are capable of solving the tasks on forming search indexes. The most common among them are the following algorithms: Soundex [6], NYSIIS [7], Daitch-Mokotoff Soundex [8], Metaphone [9], Double Metaphone [10], Russian Metaphone [4] and Polyphon [5], Caverphone [11], etc.

Calculation of the fuzzy match range based on different types of algorithms was performed in [2]. The combinations of methods that make it possible to test and check algorithms on accuracy correspond to the rules of English grammar. However, these rules cannot be fully applied to the Slavic language group, so determining the conformity range for the Ukrainian language requires a separate study. Historical grammar for genealogical research, changes in spelling or distortion of personal names constitute objective difficulties for forming universal phonetic algorithms.

Characteristics of personal names and potential sources of variations and errors are given in [3]. Although the work is an overview of the comprehensive number of frequently used methods of name alignment, but these data are only for the English language. There is an unresolved issue on the application of indicated principles for peculiarities of the Ukrainian language. And the absence of a single universal procedure indicates the necessity to develop a separate phonetic algorithm for personal Ukrainian names.

A phonetic search method that initiated the approach to error auto-correction is shown in [7]. The paper gives a set of technical solutions for automating the search process. The information system for the New York Police Service was developed. It is important to note that the hardware and software used are obsolete by now, so there are the unresolved issues related to the introduction to modern information systems. A way to overcome related difficulties could be complete transfer of a given algorithm to modern programming languages. This approach is partially applied in works [6, 8].

A modification of the phonetic algorithm for the Assamese language is given in [6]. Application of the algorithm for languages of the Indo-Aryan group indicates the universality of issues related to sound perception of information. A variant to overcome the difficulties of perception might be applying the families of phonetic algorithms for individual languages and language groups, as well as splitting the areas of application within each language.

An attempt at the universalization of phonetic search for a Slavic group of languages was reported in [8]. Although several general rules are given for search universalization, but this approach does not provide for the optimal search for each particular language. That suggests that the digitizing algorithms for individual industries with highly specialized terms require a separate revision for each language. The development of such algorithms forms a demand for a separate linguistic study of the possibilities for simplifying the groups of consonants for certain specific cases of use.

Using the Metaphone algorithm to form phonetic keys is proposed in [9, 10]. Paper [9] reports one of the first attempts to apply Metaphone for the automated search based on an outdated hardware-software base. An option to overcome the technological gap is the reengineering of the algorithm in accordance with modern hardware, as well as the refactoring of the software module code based on modern programming approaches.

A model based on two phonetic algorithms (Soundex and Metaphone) was presented in [10]. The algorithm modification makes it possible to improve algorithm operation for individual specific cases. However, there remains an unresolved issue about universal adaptation of the algorithm for other alphabet languages. A variant to overcome these difficulties is to analyze each alphabet language and form a separate set of rules. This approach was implemented for the Russian language in papers [4, 5], and [11], however, due to the differences in pronunciation in the Ukrainian and Russian languages, the application of these algorithms does not yield a significant win.

A variant of implementing the Metaphone algorithm for the Russian language was given in [4]. This algorithm includes only five phonetic rules that are inherent in the Russian language. In the Ukrainian language, the variability of pronunciation is larger, so the number of rules must be greater. That allows us to argue that universal phonetic rules cover a very limited range of tasks, and optimal search lacks special rules, separate for each language.

An example of the localization of the Polyphon phonetic algorithm for the Russian language was given in [5]. This algorithm is adapted for languages with "old" grammar and with a significant number of exceptions and archaisms. The young Ukrainian grammar does not require complicated models, so a simpler Metaphone algorithm has a potential gain in performance. Improving the Polyphon algorithm for the Ukrainian language is impractical.

An attempt to adapt several algorithms for indexing place names was made in [11]. Due to a small sample of the experimental data and a small divergence between the results from the proposed algorithms, it is difficult to give an objective estimation of the quality of operation of each separate one. The work requires detailed research using larger volumes of data. And specific phonetic search rules for place names require separate study and verification.

At the same time, applying these algorithms for the phonetic search for the highly specialized (in pronunciation and notation) words/terms, for example, in the Ukrainian language, is typically accompanied by problems associated primarily with their introduction and subsequent processing. That is, one may assert that none of the algorithms mentioned above can fully solve the task of indexing highly specialized words (in terms of their sounding) taking into consideration the rules of pronunciation, under conditions of fuzzy coincidence. This is the most characteristic in the case of data deduplication in various databases and registries. The option to overcome this issue could be to use a phonetic algorithm for comparing words, Metaphone. A given algorithm is simpler in implementation and is better suited for the "young" grammars, including Ukrainian. After a certain modification, this algorithm, by finding the distance between words at the stages of entering new data and finding words based on existing databases, would make it possible to optimize the technology of indexing highly specialized words.

There are already known implementations of the Metaphone algorithm for the English, Russian [4], and other languages, but none is known for the Ukrainian language, while the impact of phonetic search optimization on the acceleration of search processes has not been investigated.

Semantic search algorithms for last names cannot provide for a significant performance improvement, as the names act as identifiers and do not carry a separate content load.

To implement the Metaphone algorithm for the Ukrainian language, a separate set of phonetic rules for normalizing last names based on existing grammar rules [12] and reference data on Ukrainian last names [13] should be developed.

The above allows us to argue that it is advisable to conduct a research on the analysis of features of pronunciation in the Ukrainian language for highly specialized terms and the formation of a separate localized adaptation of the algorithm.

### 3. The aim and objectives of the study

The aim of this research is to develop and implement an optimization technology using phonetic search to detect a fuzzy match at data deduplication in databases. The optimization should be determined based on a decrease in the volume of search indexes compared with the full sample to the use of a phonetic algorithm. A search based on the Metaphone phonetic algorithm should be applied to reduce the number of errors when indexing Ukrainian last names and titles of medical preparations.

To accomplish the aim, the following tasks have been set:
– to investigate the frequency of using Ukrainian (by origin) last names in the territory of modern Ukraine;
– to construct a phonetic algorithm for indexes using a sample of Ukrainian last names;
– to conduct an experimental research and implement an optimization technology for the phonetic algorithm for indexes using a sample of Ukrainian last names;
– to conduct an experimental research and to implement an optimization technology for search queries related to medicinal products when two related languages mix.

### 4. Studying the frequency of using Ukrainian last names

After reducing last names to the normal form, statistics were collected on the most widespread last names. Based on these data, we compiled a list of last names whose frequency of application was over 0.3 ‰ from the total quantity of last names (more than 6,100 mentions). Table 1 gives examples of the most common last names with a frequency of use over 0.8 ‰.

Dependence of frequency $F$ (measured in ppm) of last names distribution on the number of last names can be described by a power function, derived from the diagram approximation (Fig. 1 shows the theoretical distribution by dotted line and the actual one – by continuous line): $F = 2\pi n^{-e}$, where $n$ is the number of last names with equal occurrence.

The percentage of Ukrainian last names among the most common (with a frequency of use exceeding 0.3 ‰) is 88 %, and all others belong to Russian, Belarusian, and related (Fig. 2). Therefore, the use of phonetic rules of the Ukrainian language is justified.

Fig. 2 shows that the frequency of using the last names of Ukrainian origin is prevailing.

Thus, reducing typical endings can significantly save on the length of a last name without losing its meaning.

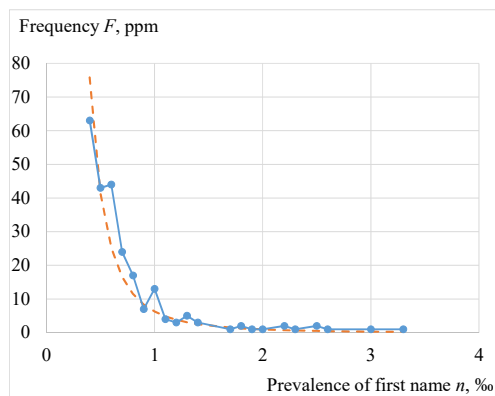

Fig. 1. Dependence of the number of last names on its occurrence

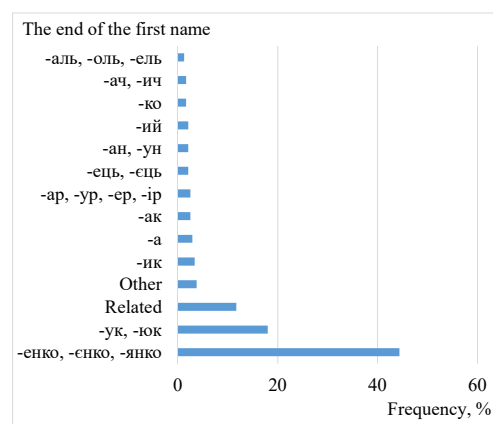

Fig. 2. Frequency of occurrence of different last names' endings

Table 1

The most common last names

| Use frequency, ‰ | Last name |
|---|---|
| 3.3 | Mel'nyk |
| 3.0 | Shevchenko |
| 2.6 | Boyko |
| 2.5 | Kovalenko, Bondarenko |
| 2.3 | Tkachenko |
| 2.2 | Koval'chuk, Kravchenko |
| 2.0 | Ivanov(a) |
| 1.9 | Oliynyk |
| 1.8 | Koval', Shevchuk |
| 1.7 | Polishchuk |
| 1.4 | Tkachuk, Bondar, Marchenko |
| 1.3 | Lysenko, Moroz, Savchenko, Rudenko, Petrenko |
| 1.2 | Kravchuk, Klymenko, Popov(a) |
| 1.1 | Pavlenko, Savchuk, Kuz'menko, Levchenko |
| 1.0 | Ponomarenko, Vasylenko, Voloshyn(a), Kharchenko, Koval'ov(a), Karpenko, Sydorenko, Havrylyuk, Mel'nychuk, Khomenko, Pavlyuk, Shvets', Popovych |
| 0.9 | Romanyuk, Chornyy(a), Panchenko, Lytvynenko, Mazur, Kushnir, Yurchenko |
| 0.8 | Dyachenko, Martynyuk, Kostyuk, Tkach, Petrov(a), Semenyuk, Prykhod'ko, Kostenko, Honcharenko, Kulyk, Kolomiyets', Bilous, Nazarenko, Volkov(a), Kravets', Kozak, Kovtun |

## 5. Constructing a phonetic algorithm for indexes based on the sample of Ukrainian last names

Given the results from our analysis, a set of phonetic rules was compiled to normalize last names based on existing grammar rules and reference data on Ukrainian last names. A given algorithm implies the following sequence of steps:

1. Remove an apostrophe (implemented at the preceding step).
2. Exclude -г-:
-ґ-→-г-.
3. Reduce vowels to their sound forms:
{-е-, -є-, -йе-, -іе-, -іо-}→-е- ("Агеєв" is dealt with at step 12);
{-а-, -я-, -іа-, -ія-}→-а-;
{-і-, -ї-, -и-}→-и-;
{-у-, -ю-}→-у-;
{-о-, -йо-}→-о- ("Йосипов").
4. Reduce non-syllabic -у ("ў") at a word's end:
-у→-в (at a word's end).
5. Change letter clusters formed from -ськ-:
{-ськ-, -сськ-}→-1-;
{-зьк-, -гськ-, -жськ-, -зськ-}→-2-;
{-цьк-, -дськ-, -тськ-, -кськ-, -чськ-, -цськ-}→-3-.
6. Remove -ь-.
7. Replace assimilated sonants (affricates and "г" are not considered due to extreme rarity):
{п-→б-, х-→г-, т-→д-, ш-→ж-, с-→з-} before {-б, -г, -д, -ж, -з}.
8. Combine crevice breath consonants:
-хв-→-ф-.
9. Replace consonants assimilated into consonant groups:
{-сч-, -жч-, -шч-, -щч-} → -щ-.
10. Simplify consonant groups:
-стн-→-сн-;
-здн-→-зн-;
-слн-→-сн-;
-стл-→-сл-;
-шчн-→-шн-.
11. Simplify others:
-цв-→-ц-.
12. Replace duplicate letters with one (both consonants and vowels).
13. Compress endings longer than 3 symbols:
-авко→-A;
{-айко, -айка}→-B;
-айло→-C;
-анко→-D;
-ашко→-E;
-евич→-F;
-евка→-G;
{-ейко, -ейка}→-H;
{-енка, -енко}→-I;
-ечко→-J;
-ешко→-K;
-ийло→-L;
-иско→-M;
-ишин→-N;
-ишко→-O;
-ович→-P;
-онко→-Q;
-очко→-R;
-уник→-S;
{-унко, -унка}→-T;
{-ушко, -ушка}→-U.

This order of rules is important, for example, the soft sign is removed only after replacing letter clusters obtained from -ськ-. When one implements the above rules, the algorithm shown in Fig. 3 is obtained.

Some rules are absorbed by others, so a database query optimization is applied to reduce execution time. An example of the phonetic algorithm implementation using regular expressions is shown in Listing 1.

*Listing 1.* Formation of indexes according to phonetic rules (with comments)

```
-- Remove -г- (step 2)
UPDATE unic SET name_mod=
replace(lastname, 'ґ', 'г');

-- Reduce vowels to their sound forms (step 3)
UPDATE unic SET name_mod=
translate(name_mod, 'іїєяю', 'ииеау');
UPDATE unic SET name_mod=
replace(name_mod, 'йе', 'е');
UPDATE unic SET name_mod=
replace(name_mod, 'іе', 'е');
UPDATE unic SET name_mod=
replace(name_mod, 'іо', 'е');
UPDATE unic SET name_mod=
replace(name_mod, 'іа', 'а');
UPDATE unic SET name_mod=
replace(name_mod, 'йо', 'о');

-- Reduce non-syllabic -у ("ў") at a word's end (step 4)
UPDATE unic SET name_mod=
regexp_replace(name_mod, 'у$', 'в', 'g');

-- Replace suffixes formed from -ськ- (step 5)
UPDATE unic SET name_mod=
regexp_replace(name_mod, '(д|т|к|ч|ц)ськ', '3', 'g');
UPDATE unic SET name_mod=
replace(name_mod, 'цьк', '3');
UPDATE unic SET name_mod=
regexp_replace(name_mod, '(г|ж|з)ськ', '2', 'g');
UPDATE unic SET name_mod=
replace(name_mod, 'зьк', '2');
UPDATE unic SET name_mod=
regexp_replace(name_mod, 'с+ьк', '1', 'g');

-- Remove -ь- (step 6)
UPDATE unic SET name_mod=
replace(name_mod, 'ь', '');

-- Replace assimilated sonants (step 7)
UPDATE unic SET name_mod=
regexp_replace(name_mod, 'п(б|г|д|ж|з)', 'б\1', 'g');
UPDATE unic SET name_mod=
regexp_replace(name_mod, 'х(б|г|д|ж|з)', 'г\1', 'g');
UPDATE unic SET name_mod=
regexp_replace(name_mod, 'т(б|г|д|ж|з)', 'д\1', 'g');
UPDATE unic SET name_mod=
regexp_replace(name_mod, 'ш(б|г|д|ж|з)', 'ж\1', 'g');
UPDATE unic SET name_mod=
regexp_replace(name_mod, 'с(б|г|д|ж|з)', 'з\1', 'g');

-- Combine crevice breath consonants (step 8)
UPDATE unic SET name_mod=
replace(name_mod, 'хв', 'ф');
```

*-- Replace consonants assimilated into consonant groups (step 9)*
**UPDATE** unic **SET** name_mod=*regexp_replace*(**name_mod**, **'(с|ж|ш|щ)ч'**, **'щ'**, **'g'**);

*-- Simplify consonant groups (step 10)*
**UPDATE** unic **SET** name_mod= *regexp_replace*(**name_mod**, **'с(т|л)н'**, **'сн'**, **'g'**);
**UPDATE** unic **SET** name_mod= *replace*(**name_mod**, **'здн'**, **'зн'**);
**UPDATE** unic **SET** name_mod= *replace*(**name_mod**, **'стл'**, **'сл'**);
**UPDATE** unic **SET** name_mod= *replace*(**name_mod**, **'шчн'**, **'шн'**);

*-- Replace others (step 11)*
**UPDATE** unic **SET** name_mod= *replace*(**name_mod**, **'цв'**, **'ц'**);

*-- Replace duplicate letters with one (step 12)*
**UPDATE** unic **SET** name_mod= *regexp_replace*(**name_mod**, **'(\w)\1+'**, **'\1'**, **'g'**);

*-- Compress endings longer than 3 symbols (step 13)*
**UPDATE** unic **SET** name_mod= *regexp_replace*(**name_mod**, **'авко$'**, **'А'**, **'g'**);
**UPDATE** unic **SET** name_mod= *regexp_replace*(**name_mod**, **'(айко|айка)$'**, **'В'**, **'g'**);
...

Due to the use of regular expressions (POSIX) and a large amount of data, initial formation of indexes can take considerable time, so the optimization of the index forming process makes it possible to save up to 15 % of CPU time. Fig. 4 shows time cost of each operation, ranked for the time of execution.

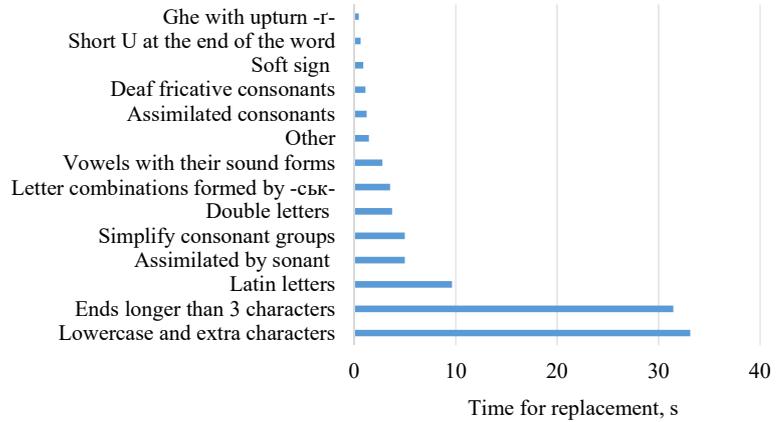

Fig. 4. Time cost diagram for different types of last name modifications

Fig. 4 shows that most resources are spent on preliminary activities to remove extra characters and reduce data to a single unified form. Because of the complexity of a regular expression with multiple nested cyclic sorting, the procedure takes three times more time than other index optimization procedures. A little less time is required to shorten the last names' endings, but nevertheless the gain in the end length is 300 %.

**6. Design of the experiment to validate a phonetic algorithm on the sample of Ukrainian last names**

For data analysis, the database of the depersonalized last names the size of 17,631,472 was used, which is approximately 41 % of the Ukrainian population according to the State Statistics Service of Ukraine for 2018 (excluding the temporarily occupied territory of the Autonomous Republic of Crimea and the city of Sevastopol) [14].

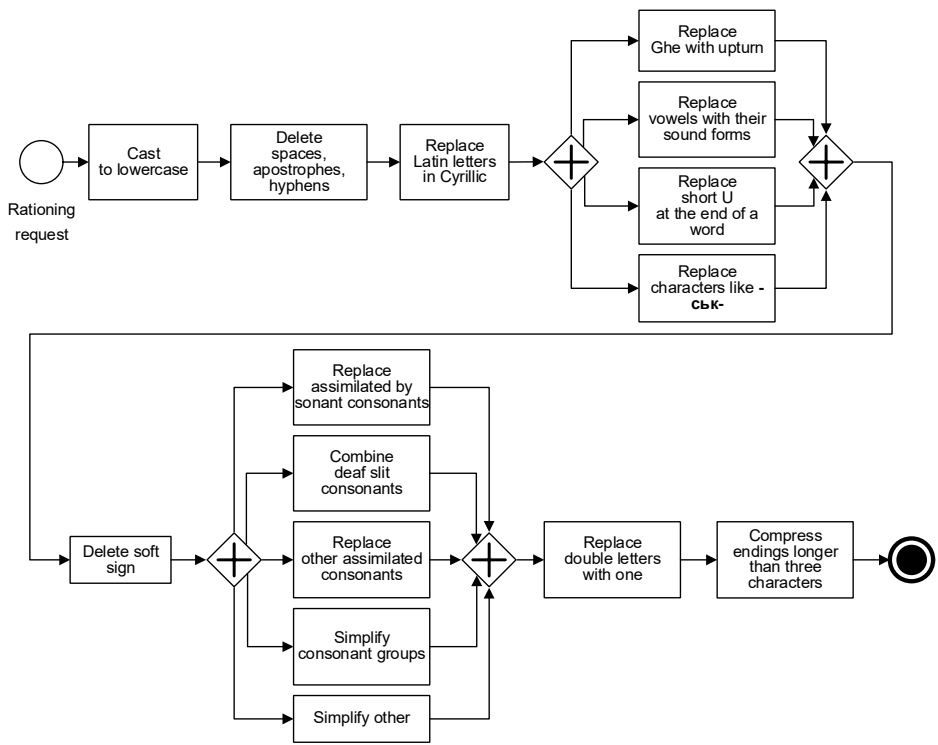

Fig. 3. BPMN diagram of the index acquisition process

All calculations and simulations were conducted in the environment of the database PostgreSQL version 10.5 [15] employing software JetBrains DataGrip version 2019.1.4 [16] on the virtual Amazon Web Services hardware at a 4×2.3 GHz processor and memory size of 16 GB. All test tables were in the first normal form (1NF).

The data are entered to the database according to simplified validation rules (for example, ban on entering numbers), therefore errors in the last names are possible. Additional reduction of data makes it possible to solve problems with 8,685 records (0.04 %). After removing all the redundant items (dashes, spaces, other service characters) and replacing the mistakenly typed ones (especially often in the beginning of the last name), it is necessary to reduce the Latin characters to Cyrillic (reducing rules are shown in Fig. 5). In addition, at the first stage, the apostrophe (Listing 2) is removed, as it carries only a partial phonetic load.

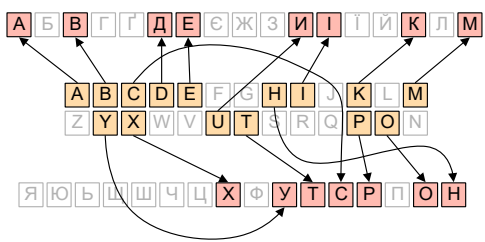

Fig. 5. Similarities in the notation of Cyrillic and Latin symbols

*Listing 2.* Creation of "pure" data

-- *Convert to lowercase, remove all spaces, and remove all double hyphens as well as hyphens at start and end of last name. Remove apostrophe*
**UPDATE** lastnames SET name_mod=
*regexp_replace*(*regexp_replace*
(*regexp_replace*(*replace*(*lower*(lastname),
' ', ''), '(-{2,})+', '-' , 'g'), '^-|-$|''$', ''),
'[^абвгґдеєжзиіїйклмнопрстуфхцчшщьюяa-z-]+', '','g');

-- *In all Latin last names, replace letters that have similar spellings (including similarity for uppercase letters)*
**UPDATE** lastnames SET name_mod=
*translate*(lastname, 'abcdehikmoptuxy',
'авсdенікмортиху')
WHERE lastname IN (SELECT lastname
FROM lastnames WHERE name_mod ~ '[a-z]');

Double and triple last names do not need to be reduced because of their uniqueness. The number of such last names is 19,290 (3.6 %). For all other last names, one can apply phonetic rules of normalization. It should be noted that the number of letters in the last names can be two or more.

### 7. Analysis of phonetic transformation results using a sample of Ukrainian last names

Productivity of index application can be validated based on two parameters:
– the ratio of indexes number to the full sequence or the average of the number of elements per index;
– a decrease in indexes volume (number of characters).
To calculate the gain, one can use optimization coefficients for the number of lines and by character volume:

$$\begin{cases} K_{opt}^{num} = \left(1 - \dfrac{I^{num}}{N^{num}}\right) \cdot 100\,\%, \\ K_{opt}^{vol} = \left(1 - \dfrac{I^{vol}}{N^{vol}}\right) \cdot 100\,\%, \end{cases} \quad (1)$$

where $I^{num}$ and $I^{vol}$ are the indexes for number and volume; $N^{num}$ and $N^{vol}$ are the complete sequences for number and volume.

Results from calculating the gain are given in Table 2.

Table 2

Gain calculation

| Sampling type | Full volume | Index volume | Optimization coefficient, % |
|---|---|---|---|
| For number, row | | | |
| Structured | 547,825 | 434,495 | 20.7 |
| Full | 17,631,472 | | 97.5 |
| For volume, symbol | | | |
| Structured | 9,213,759 | 8,358,969 | 9.3 |
| Full | 262,767,707 | | 96.8 |

Table 2 shows that a gain from different methods of calculation (based on the number of lines or symbols) is from 96.8 to 97.5 % for a full sample, which saves CPU time and RAM on the servers with databases. The saving of server computational resources is obtained by searching not in full order, but only based on indexes that form on the client side. Thus, the system employs distributed calculations, which significantly reduces the load when forming indexes. Conversion is applied only once – when adding new elements to the server database; all subsequent calculations based on a given algorithm are conducted remotely.

### 8. Design of the experiment for testing a phonetic algorithm using a sample of medicines

For some industries, there may be additional conditions, for example, when searching for medicines using the unpatented medical title and trade mark there is a need to search in two languages simultaneously: Ukrainian and Russian. However, in these related languages there are different assimilation rules for voiced and voiceless, so the rules for building indexes are enlarged and simultaneously include only joint phonetic features. Thus, the sequence of steps for medicines takes the following form:

1. Reduce to lower case.
2. Replace points, commas, hyphens, caps, other special characters (®, ™, &, *, etc.), and double spaces with spaces.
3. Replace the Latin symbols with Cyrillic (Fig. 5).
4. Assign the titles that consist of several words to separate cells.
5. Remove all titles with a length smaller than four characters.
6. Remove apostrophe, -ь- and -ъ-.
7. Exclude -г-:
-г-→-г-.
8. Reduce vowels to their sound forms:
{-е-, -є-, -э-, -йе-, -ие-, -іе-, -іо-, -ио-}→-е-;
{-а-, -я-, -іа-, -ія-, -иа-, -ия-}→-а-;
{-і-, -ї-, -и-, -ы-}→-и-;

{-у-, -ю-}→-у-;
{-о-, -йо-, -ё-}→-о-.

9. Replace duplicate letters with one (both consonants and vowels).

Testing the Morion guide [17] yielded 83,637 medicines. We separated from these data the trademarks and unpatented titles in terms of active substance in the Ukrainian and Russian languages. The result received is 133,598 words, which are reduced to 23,198 unique ones. After creating phonetic indexes based on the described algorithm, 16,049 keys were received and the optimization coefficient in line with formula (1) for the number of lines amounted to 30.8 %.

Application of a given algorithm makes it possible to automate the process of finding medicines in medical information systems both by pharmaceutical employees and users in situations when a drug is recorded by ear or illegibly written in a prescription.

### 9. Recommendations for introducing a phonetic algorithm for indexes

When introducing the algorithm for constructing phonetic indices, there is a problem of integration with existing search engine optimization systems: semantic or other universal algorithms based on the search for differences in sequences. For the case of last names, the use of semantic search does not make much sense, so there are the universal algorithms for finding differences. The simultaneous use of multiple algorithms does not result in a significant increase in the quality of search, but requires the multiplication of hardware resources (and, consequently, could lead to increased delays).

The procedure for applying phonetic algorithms is stable: first clean the data from unnecessary symbols and other "garbage" and only then get on with all transformations.

### 10. Discussion of results of studying a phonetic algorithm for indexes

Highly specialized (in pronunciation and spelling) words/terms, in particular, names of people, are typically not unique, their notation may not fall under any rules of phonetics, while their length when they are spelt in different languages may not match. With the advent of the Internet, this situation has generally become critical and could lead to the fact that multiple copies of e-mails are sent to one address. It is possible to solve the specified issue by using phonetic algorithms that compare words such as the modified Metaphone algorithm given in Listing 1. It takes into consideration, first of all, peculiarities in the formation of Ukrainian last names (Table 1) and the titles of medicines [17], and, in contrast to those existing [4, 5, 11], makes it possible to form phonetic transformations for indexes of highly specialized Ukrainian words/terms.

The results obtained in the current study have high optimization ratios derived from (1), which reach 97.5 % (Table 2). The high results of optimization are obtained through the use of the phonetic algorithm to data, structured in terms of formation and origin.

The proposed localization of the phonetic search algorithm makes it possible to optimize search exactly for the Ukrainian language, as shown in Fig. 4. In addition, an attempt at the mixed search has been made to tackle the Ukrainian and Russian languages simultaneously, which significantly expands possibilities to apply the algorithm (Listing 2) in actual information systems and makes it possible for users with a low level of literacy to enjoy search without constraints. When employing the generated indexes, one can mark an increase in productivity at the stage of entering new data and finding words based on existing databases.

This research has examined only two types of highly specialized terms. For each individual field of knowledge, a separate algorithm must be developed in order to achieve optimal algorithm performance. In addition, the results of this study could be used to implement the search in the dialects of the Ukrainian language, as well as a variety of variations in blending two related languages, for example, the so-called surzhyk, baláchka, or trasyanka.

Among the disadvantages of our study is the need to modify the algorithm for various fields of knowledge, for example, operations with last names imply a set of endings, which is inherent only for last names of Ukrainian origin.

At present, there are not known implemented other phonetic algorithms for the Ukrainian language, so it is not possible to compare performance or other parameters. In addition, the index formation rules differ even for the related grammars, so any numerical comparison based on the realization of the same algorithm in another language would not yield matching results. For different languages, test samples, as well as the sets of rules based on algorithms, are different, so it is difficult to define parameters for comparison.

In the future, we plan, first, to construct similar rules for Ukrainian names and patronymic names, and, second, to improve the algorithm for the detection of last names, which are spelt in different varieties (male and female). And to conduct a detailed research on the possibilities of simultaneous use of universal algorithms, based on the search for differences in sequences, as well as phonetic rules of index building. Application of a given method requires additional costs:

– to adapt the algorithm for a specific application area;
– to finalize databases (adding individual structures in which indexes are to be stored);
– to code (simultaneously at server and client parts);
– to check the conformity of results to user requests.

The phonetic algorithm could be applied in future to the search optimization tasks based on a last name, data deduplication processes (merging multiple accounts in one), autocomplete forms, and fill options hints. The results obtained could also be used to reduce the number of errors when filling electronic forms (questionnaires) in state and commercial information systems.

### 11. Conclusions

1. We have examined the frequency of using last names of Ukrainian origin, including their typical endings in the territory of modern Ukraine, which, when processing them, can significantly save time of search based on the length of a last name without losing its meaning. The sample amounted to approximately 41 % of the population. Due to the high percentage of Ukrainian last names among the most frequently occurring ones (88 %), we can argue that the use of phonetic search makes it possible to cover most search queries.

2. Based on the statistical data and rules of formation of Ukrainian last names, we have constructed an algorithm of

phonetic conversion for indexes for last names of Ukrainian origin. A given algorithm of phonetic conversion has made it possible to reduce the cyclical sorting procedure by three times and to get a gain of almost 300 %.

3. Theoretical results have been confirmed in the course of a field experiment using the samples of Ukrainian last names, which has allowed us to boast about saving CPU time and RAM volume at servers that host databases and optimizing search processes with an optimization coefficient of 30.8 %.

4. We have also tested algorithm performance for the mixed search for medicines in the Ukrainian and Russian languages simultaneously. The maximum gain in terms of search speed, when using phonetic indexes, reaches 97.5 %, which makes it possible to implement a given technology of optimization for the bilingual audience of users. This algorithm, along with the Levenshtein distance calculation algorithms, and other methods for finding printing errors, could significantly improve the quality of search results.

The results of this study allow us to argue that the research on the analysis of features of pronunciation in the Ukrainian language for highly specialized terms and the formation of indexes on the basis of the Metaphone algorithm corresponds to the tasks for this study and to the needs of search in modern information systems. The gain from the implementation of a given technology of optimization based on various calculation methods is from 96.8 to 97.5 %, which makes it possible to save resources of database servers and to find highly specialized terms at their fuzzy match.

### Acknowledgement

Authors express their gratitude to management team at the medical information system Helsi Ltd. [18] for providing access to depersonalized medical data, databases on medicines, and information resources, to perform analysis and construct an algorithm of the phonetic conversion for search indexes.